\documentclass[10pt,twocolumn,letterpaper]{article}

\usepackage{3dv}
\usepackage{times}
\usepackage{epsfig}
\usepackage{graphicx}
\usepackage{amsmath}
\usepackage{amssymb}
\usepackage{comment}
\usepackage{color}
\usepackage{contour}
\usepackage{xkeyval}
\usepackage{multirow}

\newcommand{\wh}{\color{white}} 
\contourlength{.5pt}
\newcommand{\figlabel}[1]{\sffamily\bfseries\wh\scriptsize\contour{black}{#1}} 
\makeatletter
\newlength{\sfp@hseplen}\newlength{\sfp@vseplen}
\define@cmdkey{subfigpos}[sfp@]{vsep}[0.6\baselineskip]{\setlength{\sfp@vseplen}{\sfp@vsep}}
\define@cmdkey{subfigpos}[sfp@]{hsep}[2.5pt]{\setlength{\sfp@hseplen}{\sfp@hsep}}
\newcommand{\subfigimg}[3][,]{\setkeys{Gin,subfigpos}{vsep,hsep,#1}\setbox1=\hbox{\includegraphics{#3}}\leavevmode\rlap{\usebox1}\rlap{\hspace*{8pt}\raisebox{\dimexpr\ht1-7pt}{\figlabel{#2}}}
  \phantom{\usebox1}
}
\makeatother

\threedvfinalcopy 

\def\threedvPaperID{40}

\ifthreedvfinal\fi
\begin{document}

\title{Cascaded Scene Flow Prediction using Semantic Segmentation}

\author{Zhile Ren\\
Brown University\\
{\tt\small ren@cs.brown.edu}
\and
Deqing Sun\\
NVIDIA\\
{\tt\small deqings@nvidia.com }
\and
Jan Kautz\\
NVIDIA\\
{\tt\small jkautz@nvidia.com }
\and
Erik B. Sudderth\\
UC Irvine\\
{\tt\small sudderth@uci.edu}
}

\maketitle

\setlength{\abovedisplayskip}{2pt plus 3pt}
\setlength{\belowdisplayskip}{2pt plus 3pt}

\begin{abstract}
Given two consecutive frames from a pair of stereo cameras, 3D scene flow methods simultaneously estimate the 3D geometry and motion of the observed scene. 
Many existing approaches use superpixels for regularization, but may predict inconsistent shapes and motions inside rigidly moving objects.
We instead assume that scenes consist of foreground objects rigidly moving in front of a static background, and use semantic cues to produce pixel-accurate scene flow estimates.
Our cascaded classification framework accurately models 3D scenes by iteratively refining semantic segmentation masks, stereo correspondences, 3D rigid motion estimates, and optical flow fields.
We evaluate our method on the challenging KITTI autonomous driving benchmark, and show that accounting for the motion of segmented vehicles leads to state-of-the-art performance.\end{abstract}

\section{Introduction}
The \emph{scene flow}~\cite{vedula1999three} is the dense 3D geometry and motion of a dynamic scene.
Given images captured by calibrated cameras at two (or more) frames, a 3D motion field can be recovered by projecting 2D motion (optical flow) estimates onto a depth map inferred via binocular stereo matching.
Scene flow algorithms have many applications, ranging from driver assistance~\cite{muller2011feature} to 3D motion capture~\cite{Furukawa:2010:Dense}.

The problems of optical flow estimation~\cite{sun2010secrets,Baker:2011:DEO} and binocular stereo reconstruction~\cite{Scharstein:2002:Taxonomy} have been widely studied in isolation. Recent scene flow methods~\cite{lv2016continuous, yamaguchi2014efficient,vogel2013piecewise} leverage 3D geometric cues to improve stereo and flow estimates, as evaluated on road scenes from the challenging KITTI scene flow benchmark~\cite{menze2015object}.
State-of-the-art scene flow algorithms~\cite{vogel20153d,menze2015object} assume superpixels are approximately planar and undergo rigid 3D motion.  Conditional random fields then provide temporal and spatial regularization for 3D motion estimates.
Those methods generally perform well on background regions of the scene,
but are significantly less accurate for moving foreground objects.

Estimating the geometry of rapidly moving foreground objects is difficult, especially near motion boundaries.  Vehicles are particularly challenging because painted surfaces have little texture, windshields are transparent, and reflections violate the brightness constancy assumptions underlying stereo and flow likelihoods.
However, accurate estimation of vehicle geometry and motion is critical for autonomous driving applications.
To improve accuracy, it is natural to design models that separately model the motion of objects and background regions~\cite{bideauECCV16,menze2015object}.

Several recent methods for the estimation of optical flow~\cite{bai2016exploiting,Hur:2016:Joint,sevilla2016optical,bideauECCV16} have used semantic cues to improve accuracy.
While motion segmentation using purely bottom-up cues is challenging,
recent advances in semantic segmentation~\cite{zhang2016instance,dai2016} make it possible to accurately segment traffic scenes given a single RGB image.
Given segmented object boundaries, object-specific 3D motion models may then be used to increase the accuracy of optical flow methods.

In this paper, we use instance-level semantic segmentations~\cite{dai2016} and piecewise-rigid scene flow estimates~\cite{vogel2013piecewise} as inputs, and integrate them via a cascade of \emph{conditional random fields} (CRFs)~\cite{lafferty01}.
We define pixel-level CRFs relating dense segmentation masks, stereo depth maps, optical flow fields, and rigid 3D motion estimates for foreground objects.
Due to the high dimensionality of these variables, we refine them iteratively using a cascaded classification model~\cite{heitz2009cascaded}, where each stage of the cascade is tuned via structural SVM learning algorithms~\cite{joachims2009cutting}.
We evaluate using previous scene flow annotations~\cite{menze2015object} of the challenging KITTI autonomous driving benchmark~\cite{Geiger:2012:We}, and improve on the state-of-the-art in two-frame scene flow estimation.
Our work demonstrates the importance of semantic cues in the recovery of the geometry and motion of 3D scenes.

\begin{figure*}[!t]
\centering
\includegraphics[width=0.93\linewidth]{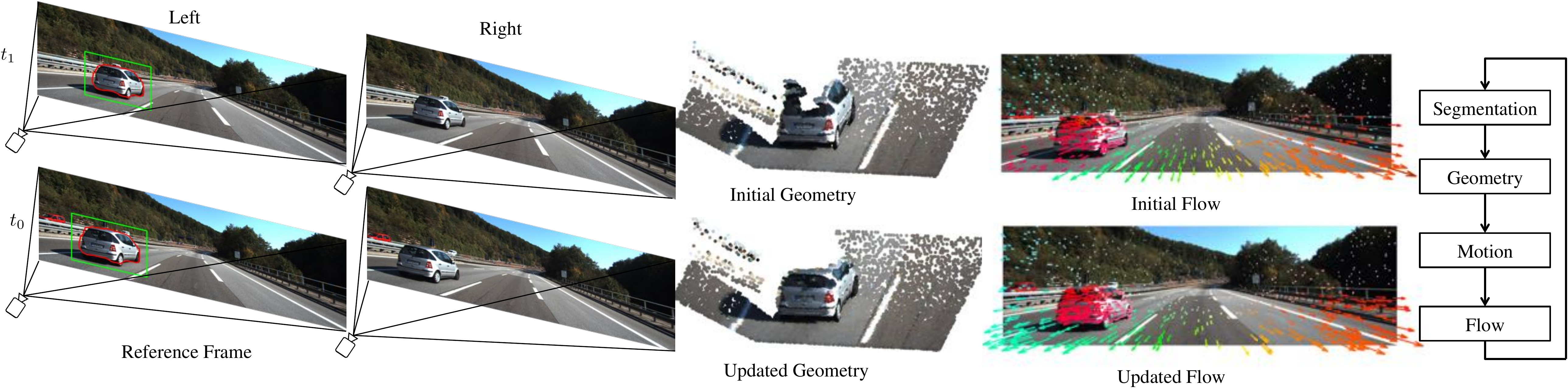}
\caption{An illustration of our method for scene flow estimation.  Given two frames from a pair of stereo cameras, and initial geometry and optical flow estimates provided by a non-semantic scene flow algorithm~\cite{vogel2013piecewise}, we use semantic segmentation cues~\cite{dai2016} to identify foreground vehicles. Our cascade of CRFs (right) then iteratively refines the inferred segmentation, geometry, 3D motion, and flow. In this example, our updated geometry estimate reduces flow errors in the windshield of the car and the adjacent road (lower left).}
  \label{fig:teaser}
\end{figure*}

\section{Related Methods for Scene Flow Estimation}
Vedula \etal~\cite{vedula1999three} first defined the scene flow as the dense 3D motion of all points in an observed scene, and recovered voxel-based flow estimates using 2D optical flow fields from several calibrated cameras. 
Huguet and Devernay~\cite{Huguet:2007:Variational} then proposed
a variational approach and jointly solved for stereo and optical flow, while  
Wedel \etal~\cite{Wedel:2008:Efficient} decoupled the stereo and flow problems for
efficiency. These classic algorithms only improve marginally over modern,
state-of-the-art stereo and optical flow methods.

Although scene flow algorithms require more input images than standard 
optical flow or stereo reconstruction methods, the task is still challenging due to the high dimensionality of the output disparity and motion fields.
To reduce the solution space,
Vogel \etal~\cite{vogel2013piecewise} introduced a \emph{piecewise rigid scene
flow} (PRSF) model and used superpixels to constrain scene flow estimation. 
For the first time, they showed that scene flow methods could outperform stereo and optical flow methods by a large margin on the challenging KITTI dataset~\cite{Geiger:2012:We}. 
In follow-up work~\cite{vogel20153d} they extended their formulation to multiple frames and improved accuracy.  However, because the PRSF model relies on bottom-up cues
for superpixel segmentation, it tends to over-segment foreground objects such as cars.
Over-segmented parts are allocated independent motion models, so global
information cannot be effectively shared.

Inspired by the success of Vogel \etal~\cite{vogel2013piecewise},
Menze and Geiger~\cite{menze2015object} annotated a new KITTI dataset with dynamic foreground objects for scene flow evaluation. 
They proposed an \emph{object scene flow} (OSF) algorithm that segments the scene into independently moving regions, and encourages the superpixels within each region 
to have similar 3D motion. 
Although the performance of OSF improved on baselines, the ``objects'' in their model are assumed to be planar and initialized via bottom-up motion estimation, so physical objects are often over-segmented.  
The inference time required for the OSF method is also significantly longer than most competing methods.

The successes of \emph{convolutional neural networks} (CNNs) for high-level vision tasks has motivated CNN-based regression methods for low-level vision.
Dosovitskiy \etal~\cite{Dosovitskiy:2015Flownet} introduced a denoising
autoencoder network, called FlowNet, for estimating optical flow.
Mayer \etal~\cite{Mayer:2015:Large} extended
the FlowNet to disparity and scene flow estimation with a large synthetic dataset.
While CNN models generate scene flow predictions rapidly, networks trained on synthetic data are not competitive with state-of-the-art methods on the real-world KITTI scene flow benchmark~\cite{menze2015object}.

Some related work integrates automatic motion segmentation with optical flow prediction~\cite{bideauECCV16,Taniai2017},
but assumes large differences between the motion of objects and cameras, and
requires multiple input frames. Exploiting the recent success of CNNs for
semantic segmentation~\cite{dai2016, zhang2016instance},
semantic cues have been shown to improve optical flow estimation~\cite{bai2016exploiting,Hur:2016:Joint,sevilla2016optical}. Concurrent work~\cite{Behl2017ICCV} also shows that semantic cues can improve scene flow estimation.
In this paper, we propose a coherent model of semantic segmentation, scene geometry, and object motion.  We use a cascaded prediction framework~\cite{heitz2009cascaded} to efficiently solve this high-dimensional inference task.
We evaluate our algorithm on the challenging KITTI dataset~\cite{menze2015object}
and show that using semantic cues leads to state-of-the-art scene flow estimates.

\section{Modeling Semantic Scene Flow}
\label{sec:sceneFlow}
Given two consecutive frames $I, J$ and their corresponding stereo pairs $I', J'$,
our goal is to estimate the segmentation mask, stereo disparity, and optical flow
for each pixel in the reference frame (Fig.~\ref{fig:teaser}).  Let
$p_i = (d_i^{(1)}, s_i^{(1)}, m_i, f_i)$ denote the variables associated with pixel~$i$ in the reference frame, where 
$d_i^{(1)} \in \mathbb{R}^+$ is its disparity, 
$s_i^{(1)} \in \{0, 1, \ldots\}$ is a semantic label ($0$ is background, positive integers are foreground object instances),
$m_i \in SE(3)$ is its 3D rigid motion (translation and rotation), and 
$f_i=[u_i, v_i]$ is its optical flow.  
We denote the disparity and semantic segmentation for each pixel in the second frame by
$q_i=(d_i^{(2)}, s_i^{(2)})$.  We only use two frames to estimate scene flow, and thus need not explicitly model motion in the second frame.

Existing scene flow algorithms make predictions at the superpixel level
without explicitly modeling the semantic content of the scene~\cite{menze2015object,vogel20153d}. 
Predictions inside each semantic object may thus be noisy or inconsistent.
In this work, we assume that the scene contains foreground objects (vehicles, for our autonomous driving application) rigidly moving across a static background.
Given an accurate semantic segmentation of some foreground object,
the geometry of the pixels within that segment should be spatially and temporally consistent, and the optical flow should be consistent with the underlying 3D rigid motion.

Due to the high dimensionality of the scene flow problem, we refine our estimates using a cascade of discriminative models~\cite{heitz2009cascaded}, with parameters learned via a structural SVM~\cite{joachims2009cutting}.  Every stage of the cascade makes a targeted improvement to one scene variable, implicitly accounting for uncertainty in the current estimates of other scene variables.
We initialize our semantic segmentation $\mathcal{S}$ using an instance-level segmentation algorithm~\cite{dai2016}, and our disparities $\mathcal{D}$ and optical flow fields $\mathcal{F}$ using the PRSF method~\cite{vogel20153d}.
We discuss their cascaded refinement next.

\subsection{Refinement of Semantic Segmentation} 
The initial single-frame segmentation is unreliable in regions with shadows and reflections. Given stereo inputs, however, our depth estimates provide a strong cue to improve the segmentation.
Therefore for each segmentation instance, we define a CRF on the pixels in its enclosing bounding box $B_i$.
We seek to estimate the foreground segmentation $s$ given an initial noisy segmentation
$\hat{s}$.

Our data term encourages the inferred segmentation $s$ to be close to the initial segmentation $\hat{s}$.
The KITTI scene flow dataset~\cite{menze2015object} generates ``ground truth'' segmentations by aligning approximate CAD models, and these annotations are often inaccurate at object boundaries, thus violating that assumption that foreground and background objects typically have distinct color and geometry.  To add robustness,
we define a feature by computing the signed distance of pixel $i$ to the 
original segmentation border and using a sigmoid function to map these distances to $[0, 1]$, denoted by $\phi_{\text{dist}}(i, \hat{s})$.
The data energy for our CRF model is then
\begin{equation}
\begin{aligned}
  E_{\text{seg}}^{\text{data}}(\mathcal{S}) & \!=\! \sum_{i\in B_i} \Big[\lambda_1 \!+\! \lambda_2 \phi_{\text{dist}}(i, \hat{s})\Big]\delta(s_i\!=\!0, \hat{s}_i\!=\!1)\\
                                            &\quad + \Big[\lambda_3 \!+\! \lambda_4 \phi_{\text{dist}}(i, \hat{s})\Big]\delta(s_i\!=\!1,\hat{s}_i\!=\!0).
\end{aligned}
\label{eq:EsegData}
\end{equation}

\begin{figure}[!t]
\centering
\begin{tabular}{cc}
\includegraphics[width=0.48\linewidth]{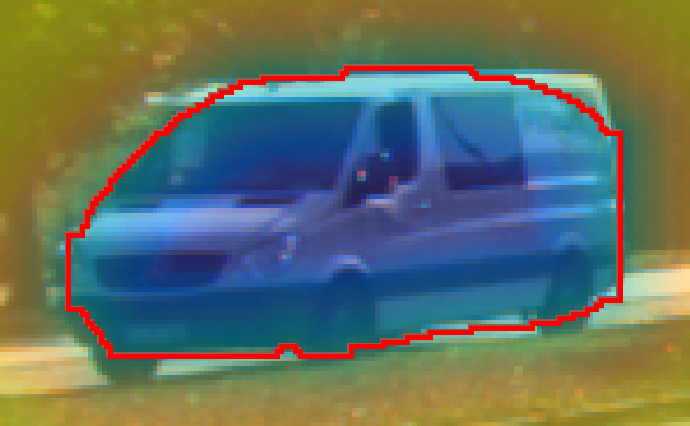} &
\includegraphics[width=0.48\linewidth]{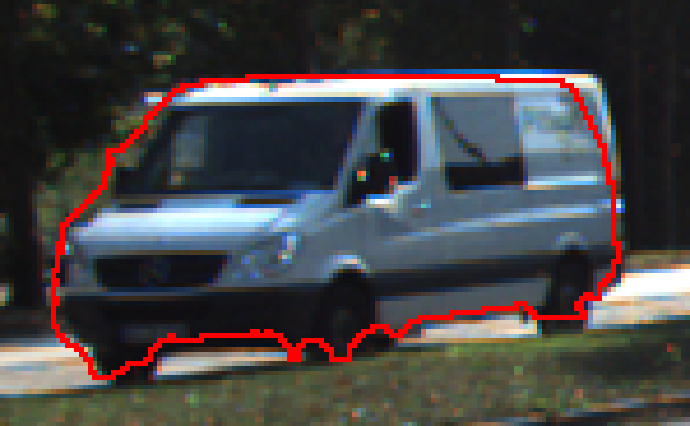}\\
$\text{\small Signed Distance }\phi_{\text{dist}}(\cdot,\hat{s})$ & {\small ``Ground Truth'' Segmentation} \\
\includegraphics[width=0.48\linewidth]{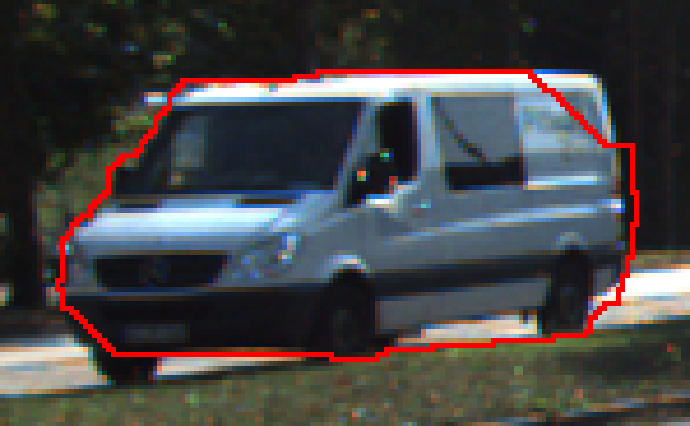} &
\includegraphics[width=0.48\linewidth]{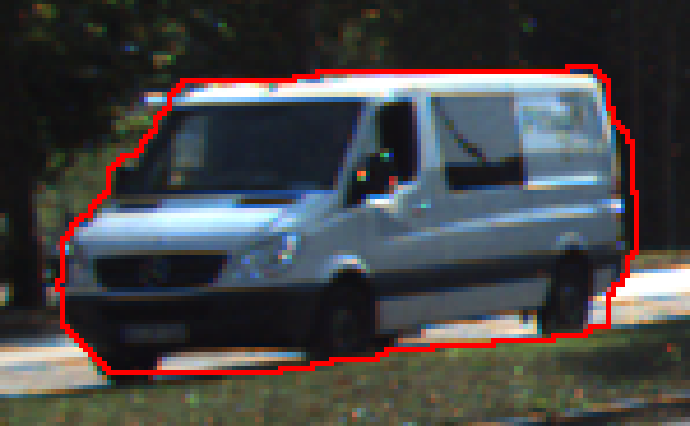}\\
{\small CRF Excluding }$\phi_{\text{dist}}$ & {\small CRF Including }$\phi_{\text{dist}}$
  \end{tabular}
  \vspace*{3pt}
  \caption{The ``true'' KITTI segmentations~\cite{menze2015object} are approximate (top right).  By incorporating a signed distance feature $\phi_{\text{dist}}(i,\hat{s})$ (top left), CRF segmentation accuracy improves (bottom).}
  \label{fig:seg_feature}
\end{figure}

\noindent We demonstrate the benefits of our signed distance feature $\phi_{\text{dist}}(i, \hat{s})$ in Fig.~\ref{fig:seg_feature}.
By allowing the CRF to reduce confidence in $\hat{s}$ near boundaries, this feature allows other image-based cues to improve segmentation accuracy.

To allow spatial regularization, we add edges $\mathcal{E}$ to our CRF connecting each pixel to its 8 spatial neighbors:
\begin{equation}
\begin{aligned}
E_{\text{seg}}^{\text{space}}(\mathcal{S}) &= \sum_{(i,j) \in \mathcal{E}}\Big[\lambda_5 + \lambda_6 \rho_{\text{img}}(I_i, I_j)  \\
&+ \lambda_7 \rho_{\text{disp}}(d_i, d_j)\Big]\delta(s_i \neq s_j).
\end{aligned}
\label{eq:EsegSpatial}
\end{equation}
Here, $\rho_{\text{img}}(I_i, I_j) = \exp\{-\frac{||I_i - I_j||}{\sigma_{\text{img}}}\}$ measures RGB color similarity,
and $\rho_{\text{disp}}(d_i, d_j)=\exp\{-\frac{|d_i - d_j|}{\sigma_{\text{disp}}}\}$ measures similarity of the current (approximate) disparity estimates.

To learn the parameters $\lambda = [\lambda_1, \ldots, \lambda_{7}]$, we use a structured SVM~\cite{joachims2009cutting}
with loss equal to the average label error within bounding box $B_i$~\cite{finley2008training}.
Feature bandwidths $\sigma_{\text{img}},\sigma_{\text{disp}}$ are tuned using validation data.
To perform inference on $E_{\text{seg}}^{\text{data}} + E_{\text{seg}}^{\text{space}}$,
we use an efficient implementation of
tree-reweighted belief propagation~\cite{wainwright2005map,kolmogorov2006convergent}.
Because pixel labels are binary, inference takes less than 0.5s.
To apply our CRF model to the scene flow problem, we independently estimate the segmentation for each instance and frame.

\subsection{Estimation of Scene Geometry}
\label{sec:stereo}
\begin{figure}[!t]
\centering
\begin{tabular}{cc}
\includegraphics[width=0.38\linewidth]{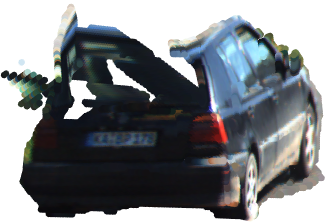} &
\includegraphics[width=0.42\linewidth]{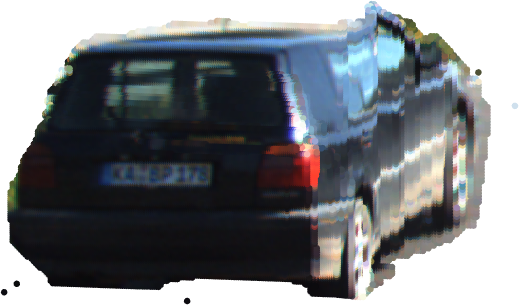}\\
{\small Initial Point Cloud~\cite{vogel2013piecewise}} & {\small Refined Point Cloud}\\
\includegraphics[width=0.48\linewidth]{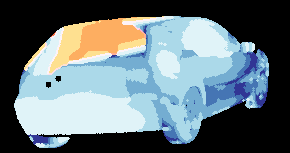} &
\includegraphics[width=0.48\linewidth]{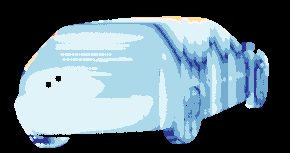}\\
{\small Initial Disparity Error} & {\small Refined Disparity Error}
  \end{tabular}
  \vspace*{2pt}
\caption{3D point clouds (top) and corresponding disparity errors (blue small, orange large) for the initial PRSF depth estimates~\cite{vogel2013piecewise}, and the refined depth estimates produced by our CRF model.}
  \label{fig:disp_exp}
\end{figure}

Given a disparity map $\mathcal{D}$ and camera calibration parameters, a 3D point cloud representation of the scene may be constructed.  Standard stereo estimation algorithms ignore semantic cues, and often perform poorly on surfaces that are shadowed, reflective, or transparent. As illustrated in Fig.~\ref{fig:disp_exp}, for autonomous driving applications the depth estimates for vehicle windshields are especially poor.
Because inaccurate depth estimates lead to poor motion and flow estimates, we design a model that enforces local smoothness of depths within inferred segmentation masks.

We define a CRF model of the pixels within each semantic segment previously inferred by our cascaded model.
For each pixel $i$ in the left camera with disparity hypothesis $d_i$, we denote its corresponding pixel in
the right camera as $P_d(i, {d_i})$. The data term is defined to
penalize the difference in smooth census transform between pixel $i$ and $P_d(i, {d_i})$:
\begin{equation}
\begin{aligned}
&E_{\text{geom}}^{\text{data}}(\mathcal{D}) = \sum_{\{i|s_i=s\}}\rho_{\text{CSAD}}(I_i, I'_{P_d(i, {d_i})}).
\end{aligned}
\label{eq:EgeomData}
\end{equation}
Here, $\rho_{\text{CSAD}}(.,.)$ is the CSAD cost~\cite{vogel2013evaluation}
for matched pixels in different images. The CSAD difference is a
convex approximation of the census transform~\cite{zabih1994non} that gives reliable pixel correspondences for many datasets~\cite{vogel2013evaluation}.

We encourage piecewise-smooth depth maps by penalizing the absolute difference of neighboring pixel depths: 
\begin{equation}
\begin{aligned}
  &E_{\text{geom}}^{\text{space}}(\mathcal{D}) = \tau_1
\sum_{(i, j) \in\mathcal{E}_s}\rho_{\text{depth}}(d_i, d_j).
\end{aligned}
\label{eq:EgeomSpace}
\end{equation}
Here $\mathcal{E}_s$ contains neighboring pixels within segment $s$, $\rho_{\text{depth}}(d_i, d_j) = |\frac{C}{d_i} - \frac{C}{d_j}|$, and $C$ is a
camera-specific constant that transforms disparity $d$ into depth $\frac{C}{d}$. We enforce consistency of pixel depths because the scale of disparities varies widely with the distance of objects from the camera.

If naively applied to the full image, simple CRF models are often inaccurate at object boundaries~\cite{vogel2013piecewise}. However as illustrated in Fig.~\ref{fig:disp_exp}, although our stereo CRF uses standard features,  it is effective at resolving uncertainties in challenging regions of foreground objects and it is much better able to capture depth variations within a single object.
Moreover, because our pairwise distances depend only on the absolute value of depth differences, 
distance transforms~\cite{felzenszwalb2006efficient} may be used for efficient inference in minimizing $E_{\text{geom}}^{\text{data}} + E_{\text{geom}}^{\text{space}}$.
On average, it takes less than 5s to perform inference in a $200\times200$ region
with 200 disparity candidates.  We refine the disparities for each frame independently.

\subsection{Estimation of 3D Motion}
If the segmentation mask and disparity estimates for each object instance were perfect, we could apply 3D rigid motion to the 3D point cloud for each segment, and project back to the image plane to recover the 2D optical flow. We let $(x_i,y_i)$ denote the \emph{motion flow} constructed in this way.
Although our imperfect geometry estimates will cause the motion flow to differ from the true optical flow $(u_i,v_i)$,
each still provides valuable cues for the estimation of the other.

\begin{figure}[!t]
\centering
\begin{tabular}{cc}
\subfigimg[width=0.48\linewidth]{Ground Truth Optical Flow}{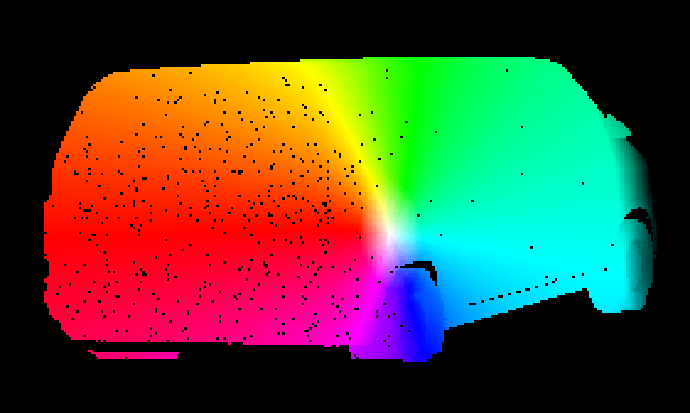} &
\subfigimg[width=0.48\linewidth]{Image $I$}{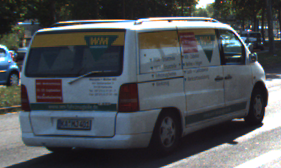}\\
\subfigimg[width=0.48\linewidth]{Initial Flow}{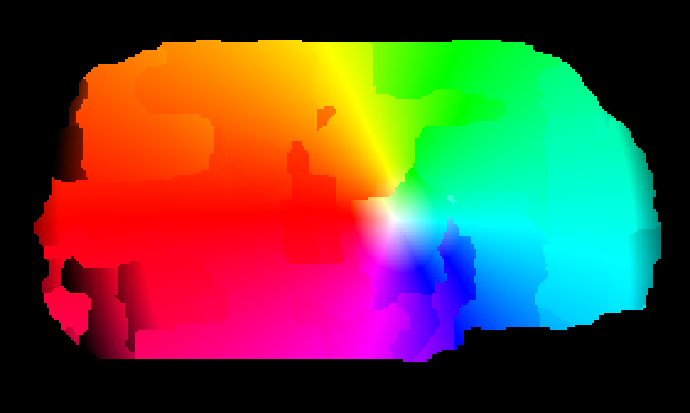} &
\subfigimg[width=0.48\linewidth]{Initial Flow Error}{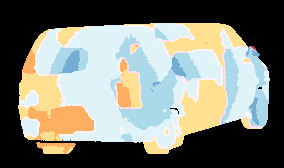}\\
\subfigimg[width=0.48\linewidth]{Motion Flow}{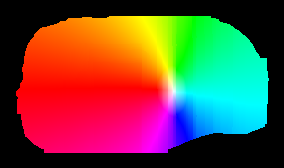} &
\subfigimg[width=0.48\linewidth]{Motion Flow Error}{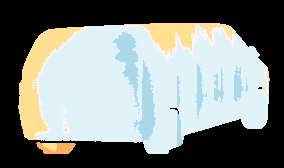}\\
\subfigimg[width=0.48\linewidth]{Refined Flow}{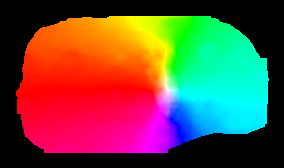} &
\subfigimg[width=0.48\linewidth]{Refined Flow Error}{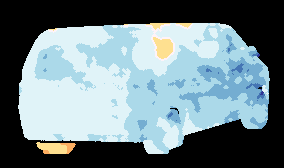}
  \end{tabular}
  \vspace*{2pt}
  \caption{Visualization of estimated flow fields (left, hue encodes orientation~\cite{sun2010secrets}) and their error (right, blue small, orange large).  A rigid 3D motion flow captures the dominant object motion, and the refined estimates from our CRF model further improve accuracy.}
  \label{fig:motion_exp}
\end{figure}

For each detected segment, we let $M=(R,t)$ denote its 3D relative motion between the first and second frames.  The motion $M$ has 6 degress of freedom: $t$ is a translation vector, and $R = (\alpha,\beta,\gamma)$ is a rotation represented by three axis-aligned rotation angles. 
We match the rigid motion $M$ to the current flow field estimate $(u,v)$ by minimizing the following energy function: 
\begin{multline}
  E_{\text{motion}}(M) = \nu (\rho(\alpha) + \rho(\beta) + \rho(\gamma))\\
    + \sum_{\{i|s_i=s\}} |x_{i}(M,d_i) - u_i| + |y_{i}(M,d_i) - v_i|.
\label{eq:rigid_motion}
\end{multline}
where $(x_i(M,d_i),y_i(M,d_i))$ is the motion flow computed from disparity $d_i$, 3D motion $M$, and the camera calibration.
We let $\rho(a)=\sqrt{a^2 + \epsilon^2}$ be the Charbonnier penalty, a smooth function similar to the $L_1$ penalty that provides effective regularization for motion estimation tasks~\cite{sun2010secrets}. We regularize $R$ to avoid unrealistically large rotation estimates.
We set the regularization constant $\nu$ using validation data, and use gradient descent to find the optimal value for $M$.  We visualize an example motion flow map in 
Fig.~\ref{fig:motion_exp}.

\subsection{Estimation of 2D Optical Flow}
The estimated motion flow from the previous stage provides valuable
cues for optical flow estimation. As in the example in Fig.~\ref{fig:motion_exp},
motion flow errors are primarily caused by imperfect geometries (or disparities). 
We thus seek a flow field $f_i=(u_i, v_i)$ such that the corresponding pixel $P_f(i, f_i)$ in the next frame matches pixel $i$, and $f_i$ does not deviate too much from $(x_i, y_i)$.
We define a CRF model of the pixels within segment $s$ in frame 1, with likelihood
\begin{equation}
\begin{aligned}
    E_{\text{flow}}^{\text{data}}(\mathcal{F}) & = \sum_{\{i|s_i=s\}} \rho_{\text{CSAD}}(I_i, J_{P_f(i, f_i)}) \\
    &+ \eta_1 (|u_i - x_{i}| + |v_i - y_{i}|).
\end{aligned}
\end{equation}
\noindent We also encourage spatially smooth flow field estimates:
\begin{equation}
\begin{aligned}
    E_{\text{flow}}^{\text{space}}(\mathcal{F}) & = \sum_{(i, j)\in \mathcal{E}_s}
    \eta_2 (|u_i - u_{j}| + |v_i - v_{j}|).
\end{aligned}
\end{equation}

\noindent While many optical flow methods use superpixel approximations to make inference more efficient~\cite{sevilla2016optical},
max-product belief propagation can be efficiently implemented for our pixel-level CRF using distance transforms~\cite{felzenszwalb2006efficient,chen2016full}.
As shown in Fig.~\ref{fig:motion_exp}, our refined optical flow improves the initial flow by smoothly varying across the segment, while simultaneously capturing details that are missed by the motion flow.

To limit the memory consumption of our optical flow algorithm, we perform inference on a down-sampled image and then use the EpicFlow~\cite{revaud2015} algorithm to interpolate back to the full image resolution.  Other recent optical flow algorithms have used a similar approximation~\cite{chen2016full,bai2016exploiting}.

\begin{figure}[!t]
\centering
\begin{tabular}{cc}
\subfigimg[width=0.45\linewidth]{image $I$}{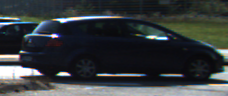} &
\subfigimg[width=0.45\linewidth]{image $J$}{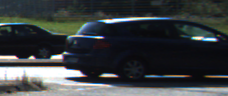}\\ 
\subfigimg[width=0.45\linewidth]{Estimated Flow}{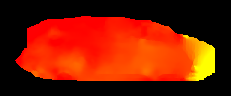} &
\subfigimg[width=0.45\linewidth]{Estimated Flow Error}{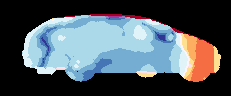}\\
\subfigimg[width=0.45\linewidth]{Motion Flow}{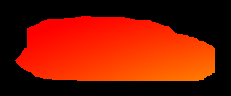} &
\subfigimg[width=0.45\linewidth]{Motion Flow Error}{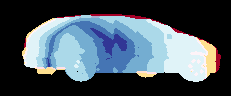}\\
\subfigimg[width=0.45\linewidth]{Fused Flow}{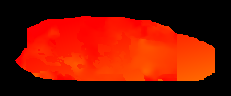} &
\subfigimg[width=0.45\linewidth]{Fused Flow Error}{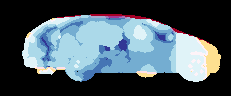}
  \end{tabular}
  \caption{Visualization of our flow fusion CRF to reduce motion errors (blue small, orange large) for out-of-border pixels.}
  \label{fig:flow_fusion}
\end{figure}

\paragraph{Motion Estimation for Out-of-Frame Pixels} 
\vspace{-12px}
We notice that the EpicFlow interpolation tends to produce significant errors for
pixels that move outside of the image border.  Outside of the camera's field of view, optical flow can only be predicted using the known 3D rigid motion, and we thus propose a flow fusion CRF~\cite{lempitsky2008fusionflow} to combine the estimated optical flow and motion flow for partially occluded objects.

In particular, we use a binary CRF to determine whether the optical flow $(u_i, v_i)$ or motion flow $(x_i, y_i)$ provides a better estimate of the true flow $(U_i,V_i)$ for each pixel $i$.
Intuitively, for within-border pixels we should use the matching cost to compare flow fields, while out-of-border pixels should be biased towards the motion flow interpolation:
\begin{multline*}
  E_{\text{fuse}}^{\text{data}}(\mathcal{F}) = 
  \omega_1 (|U_i - x_{i}| + |V_i - y_{i}|)\delta[P_f(i, f_i)\text{\small\, outside}]\\
  +\!\!\sum_{f=\{(u, v), (x, y)\}}  \sum_{\{i|s_i=s\}}\!\!\! \rho_{\text{CSAD}}(I_i, J_{P_f(i, f_i)})\delta[P_f(i, f_i)\text{\small\, inside}]. 
\end{multline*}

Spatial smoothness is encouraged for neighboring pixels:
\begin{equation}
E_{\text{fuse}}^{\text{space}}(\mathcal{F})  = \sum_{(i, j)\in \mathcal{E}}
    \omega_2 (|U_i - U_j| + |V_i - V_j|).
\end{equation}
We tune parameters $\omega_1, \omega_2$ using validation data, and minimize the energy using
tree-reweighted belief propagation~\cite{kolmogorov2006convergent}. We show in Fig.~\ref{fig:flow_fusion}
that the fused flow estimate retains many details of the optical flow, while using the motion flow to better interpolate in occluded regions.
We also apply our flow fusion technique to update the noisy background flow predictions.  See Fig.~\ref{fig:vis_results} for additional examples of our final optical flow estimates.

\section{Cascaded Scene Flow Prediction}
\begin{figure}[!t]
\centering
\includegraphics[width=0.6\linewidth]{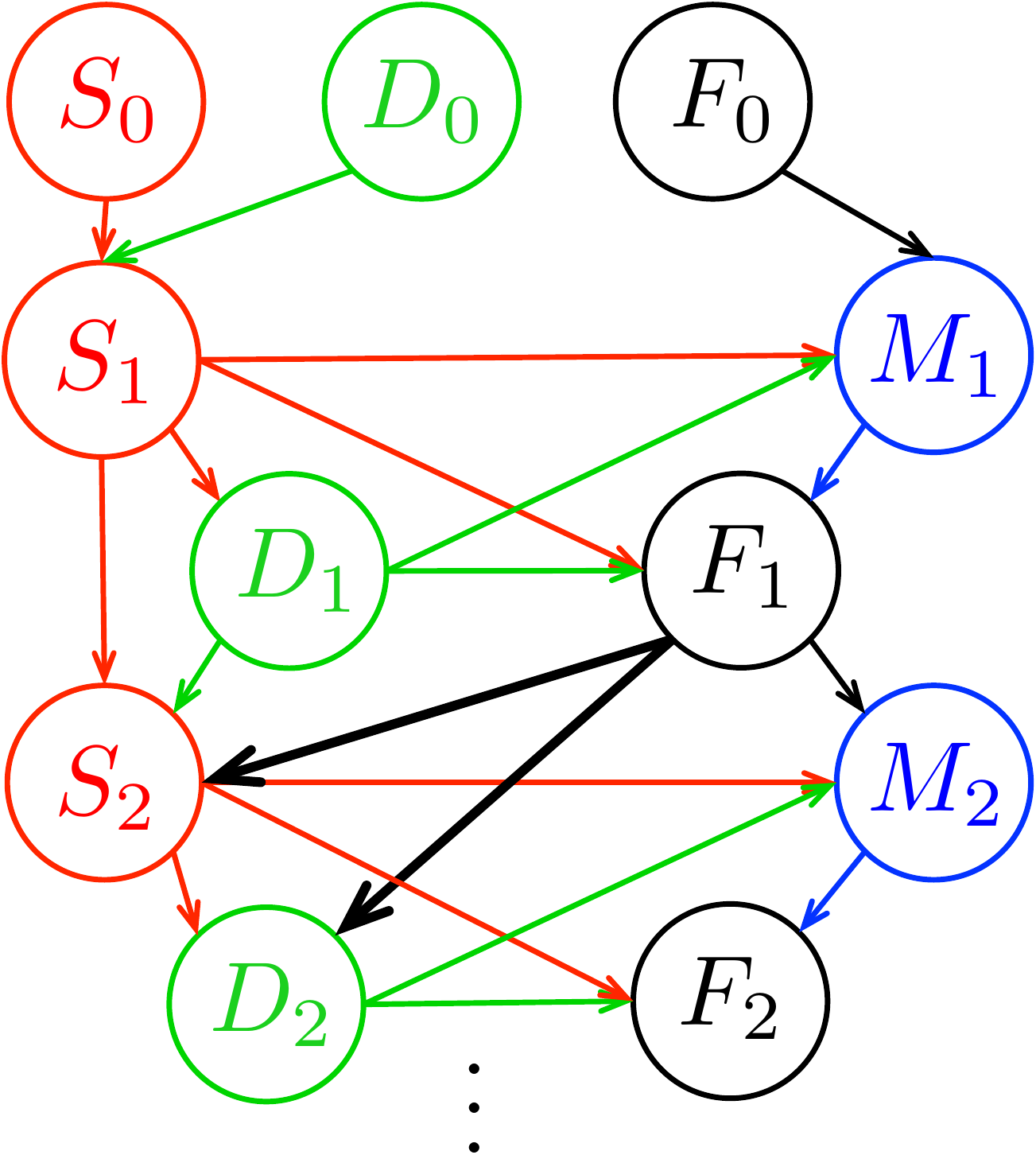}
 \caption{A directed graph summarizing our cascaded approach to the estimation of object segmentations $S$, disparities $D$, 3D rigid motions $M$, and optical flow $F$. Subscripts indicate different stages of the cascade.  The \textbf{bold arrows} represent additional temporal dependencies added for stages two and later.}
  \label{fig:cascade}
\end{figure}

The CRF models defined in Sec.~\ref{sec:sceneFlow} refine the various components of our scene model greedily, by estimating each one given the current best estimates for all others.
However, this approach does not fully utilize the temporal relationships between the segmentation and geometry at different frames.  
Also, when the initial optical flow contains major errors, our motion flow estimates will be inaccurate.
To better capture the full set of geometric and temporal relationships, we thus use multiple stages of cascaded prediction~\cite{heitz2009cascaded} to further refine our scene flow estimates.
The inputs and outputs for each stage of our cascade are summarized by the directed graph in Fig.~\ref{fig:cascade}.

\vspace*{-10pt}
\paragraph{Temporal Segmentation Consistency}
Rather than segmenting each video frame independently, in the second stage of our cascade, we use the inferred flow field $f$ to encourage temporal consistency.
Each pixel $i$ in frame 1 is linked to matched pixel $P_f(i, f_i)$ in frame 2:
\begin{multline*}
E_{\text{seg}}^{\text{time}}(\mathcal{S}) =
  \lambda_8 \delta(s^{(1)}_i = 0, s^{(2)}_{P_f(i, f_i)} = 1)\\
  \hspace*{-.25in}+ \lambda_9 \delta(s^{(1)}_i = 1, s^{(2)}_{P_f(i, f_i)} = 0)\\ 
  + \sum_{i}\Big[\lambda_{10} + \lambda_{11} \rho_{\text{CSAD}}(I_i, J_{P_f(i, f_i)})\Big]\delta(s^{(1)}_i = s^{(2)}_{P_f(i, f_i)}).  
\label{eq:EsegTemporal}
\end{multline*}
We again use S-SVM learning of CRF parameters $\lambda$ on $E_{\text{seg}}^{\text{data}}+E_{\text{seg}}^{\text{space}}+E_{\text{seg}}^{\text{time}}$, and infer segmentations using tree-reweighted belief propagation.

\begin{figure}[t]
\begin{tabular}{cc}
\subfigimg[width=0.47\linewidth]{Stereo Input}{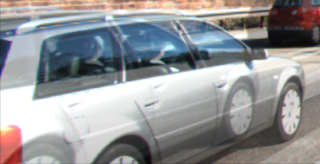} &
\subfigimg[width=0.47\linewidth]{Motion Input}{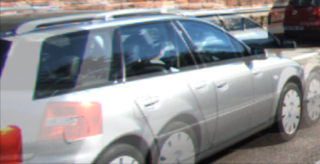} \\
\subfigimg[width=0.47\linewidth]{Initial Disparity}{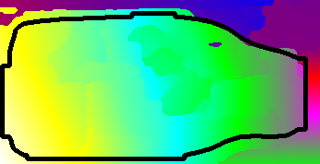} &
\subfigimg[width=0.47\linewidth]{Initial Flow}{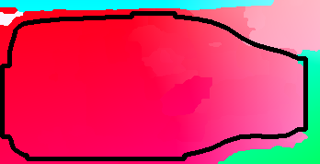} \\
\subfigimg[width=0.47\linewidth]{Iteration 1}{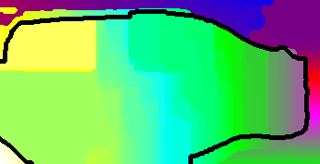} &
\subfigimg[width=0.47\linewidth]{Iteration 1}{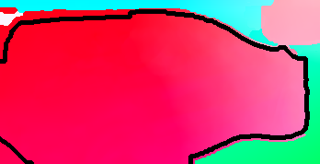} \\
\subfigimg[width=0.47\linewidth]{Iteration 2}{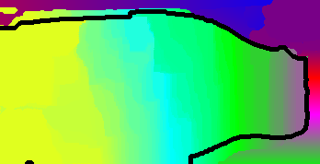} &
\subfigimg[width=0.47\linewidth]{Iteration 2}{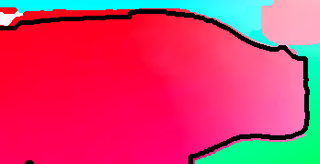} \\
\end{tabular}
\caption{From top to bottom, we visualize input frames, initial disparity 
(left) and flow (right) predictions, and the refined disparity and flow after the first and second stages of the cascade.  Our refined flow estimates from stage 1 (note object boundaries) lead to improved stereo estimates in stage 2 (upper left).}
\label{fig:cascade_exp}
\end{figure}

\paragraph{Temporal Geometric Consistency}
\vspace*{-10pt}
As in our temporal segmentation model, we also extend the stereo CRF of Sec.~\ref{sec:stereo} to encourage smooth changes for the depths of pixels linked by our optical flow estimates:
\begin{equation}
E_{\text{geom}}^{\text{time}}(\mathcal{D}) = \tau_2 \sum_{\{i|s_i^{(1)}=s\}}\rho_{\text{depth}}(d_i(m_i), d_{P_f(i, f_i)}).
\label{eq:EgeomTime}
\end{equation}
\noindent Here, $d_i(m_i)$ denotes the disparity value of pixel $i$ in the second frame when rigid motion $m_i$ is applied.  The parameters $\tau$ are learned using validation data, and efficient distance transformation~\cite{felzenszwalb2006efficient} is also used to solve $E_{\text{geom}}^{\text{data}}+E_{\text{geom}}^{\text{space}}+E_{\text{geom}}^{\text{time}}$.
Fig.~\ref{fig:cascade_exp} shows an example of the improved disparity and flow estimates produced across multiple stages of our cascade.

\paragraph{Recovery from a Poor Optical Flow Initialization}
\vspace*{-10pt}
If the initial noisy optical flow is very inaccurate, our cascade cannot recover the correct 3D motions of objects because we assume motion flow should match optical flow. 
Since our updated semantic segmentation masks $s^{(1)}$ and $s^{(2)}$ are typically very accurate,
when applying rigid motion $M$ to pixels in $s^{(1)}$, 
the shape of the new segmentation mask $s(M)$ should be similar to $s^{(2)}$.
We measure this similarity via a cost defined on the second-frame bounding box $B$:
\begin{equation}
 \frac{1}{|B|}\sum_{i\in B} \alpha S(M)_i \cdot C(S^{(2)}_i) + (1-\alpha) C(S(M)_i) \cdot S^{(2)}_i.
 \label{eq:silhouette}
\end{equation}
Here, $C(\cdot)$ is the Chamfer difference map and $\alpha=0.5$. 
This cost function is widely 
used for human pose estimation~\cite{balan2007detailed}.
If this cost exceeds 0.5, we replace the first
term in Eq.~\eqref{eq:rigid_motion} with this silhouette cost. By optimizing this modified objective in Eq.~\eqref{eq:silhouette} using standard gradient-descent, we can recover from bad motion estimates. An illustration is in the supplementary material.

\paragraph{Second Frame Disparities} 
\vspace*{-16pt}
For the KITTI scene flow dataset~\cite{menze2015object}, the ground truth 
disparity for the second frame is represented as per-pixel disparity changes with respect to the first frame. 
To predict this quantity for evaluation, we apply our estimated 3D rigid motion for each pixel to its estimated geometry in the first frame.  
The accuracy of these disparity estimates is thus strongly dependent on the performance of our motion estimation algorithm.

\paragraph{Global Energy Function}
\vspace*{-12pt}
The global energy function implicitly minimized by our cascade of CRFs can be constructed by adding all energy terms together. Our iterative optimization of subsets of variables (as in Fig.~\ref{fig:cascade}) can be seen as block coordinate descent, where the cascaded prediction framework refines the energy function to reflect the typical accuracy of previous stages.  This cascaded framework enables efficient, adaptive discretization of a large state space for flow and disparity, and is a principled way of optimizing a limited number of inference iterations~\cite{heitz2009cascaded}.

\section{Experiments}
\setlength\tabcolsep{4.5pt}
\begin{table*}
\centering
\begin{tabular}{c |  c  c  c  c  c  c  c  c  c  c  c  c c}
{\bf } &  {\bf D1-bg} & {\bf D1-fg} & {\bf D1-all} & {\bf D2-bg} & {\bf D2-fg} & {\bf D2-all} & {\bf Fl-bg} & {\bf Fl-fg} & {\bf Fl-all} & {\bf SF-bg} & {\bf SF-fg} & {\bf SF-all} &{\bf Time} \\ \hline
 PRSF  & 4.74 & 13.74 & 6.24 & 11.14& 20.47 & 12.69 & 11.73 & 27.73 & 14.39& 13.49 & 31.22 & 16.44 & 2.5min\\
 CSF   & 4.57 & 13.04 & 5.98 & 7.92 & 20.76 & 10.06 & 10.40 & 30.33 & 13.71& 12.21 & 33.21 & 15.71 & {1.3min}\\
 OSF   & 4.54 & 12.03 & 5.79 & 5.45 & 19.41 & 7.77  & 5.62  & 22.17 & 8.37 & 7.01  & 26.34 & 10.23 & 50min \\
 \textbf{SSF-O} & 4.30 & {8.72}  & 5.03 & 5.13 & {15.27} & {6.82}  & {5.42}  & 17.24 & 7.39 & {6.95}  & 25.78 & 10.08 & 52.5min \\
 \textbf{SSF-P} & {3.55} & 8.75  & {4.42} & {4.94} & 17.48 & 7.02  & 5.63  & {14.71} & {7.14} & 7.18  & {24.58} & {10.07} & 5min \\
 {ISF} & 4.12 & 6.17  & 4.46 & 4.88 & 11.34 & 5.95  & 5.40  & {10.29} & {6.22} & 6.58  & {15.63} & {8.08} & 10min \\
 \hline \hline
 OSFTC & 4.11 & 9.64  & 5.03 & 5.18 & 15.12 & 6.84  & 5.76  & 16.61 & 7.57 & 7.08 & {22.55} & 9.65 & 50min    \\
 PRSM   & {3.02} & 10.52 & {4.27} & 5.13 & {15.11} & {6.79}  & {5.33}  & 17.02 & 7.28 & {6.61} & 23.60 & {9.44} & 10min
 \end{tabular}
\caption{Scene flow results on all pixels for the KITTI test set. Under most evaluation metrics,
our SSF algorithm outperforms baseline two-frame scene flow methods such as PRSF~\cite{vogel2013piecewise}, CSF~\cite{lv2016continuous}, and
OSF~\cite{menze2015object}. OSFTC and PRSM take multiple frames as input. ISF~\cite{Behl2017ICCV} is concurrent work (to appear at ICCV 2017) that benefits from additional training data and corrected KITTI training labels.}
\label{table:sf_all_results}
\end{table*}

\setlength\tabcolsep{6pt} 
\begin{table*}
\centering
\begin{tabular}{ c | c  c  c  c  c  c  c  c  c  c  c  c }
{\bf } & {\bf D1-bg} & {\bf D1-fg} & {\bf D1-all} & {\bf D2-bg} & {\bf D2-fg} & {\bf D2-all} & {\bf Fl-bg} & {\bf Fl-fg} & {\bf Fl-all} & {\bf SF-bg} & {\bf SF-fg} & {\bf SF-all} \\ \hline
CSF    & 4.03 & 11.82 & 5.32 & 6.39 & 16.75 & 8.25 & 8.72 & 26.98 & 12.03& 10.26& 28.68 & 13.56 \\
PRSF   & 4.41 & 13.09 & 5.84 & 6.35 & 16.12 & 8.10 & 6.94 & 23.64 & 9.97 & 8.35 & 26.08 & 11.53 \\
OSF    & 4.14 & 11.12 & 5.29 & 4.49 & 16.33 & 6.61 & 4.21 & 18.65 & 6.83 & 5.52 & 22.31 & 8.52 \\
\textbf{SSF-O}  & 3.98 & 7.82 & 4.62 & 4.26 &  {12.31} & {5.70} & {4.04} & 13.18 & 5.70 & {5.44} & 21.11 & {8.25} \\
\textbf{SSF-P}  & {3.30} & {7.74}  & {4.03} & {4.12} & 14.57 & 5.99 & 4.20 & {10.81} & {5.40} & 5.70 & {19.93} & {8.25}\\
{ISF}  & {3.74} & {5.46}  & {4.02} & {4.06} & 9.04 & 4.95 & 4.21 & {6.83} & {4.69} & 5.31 & {11.65} & {6.45}\\
\hline \hline
PRSM   & {2.93} & 10.00 & 4.10 & 4.13 & 12.85 & 5.69 & 4.33 & 14.15 & 6.11 & 5.54 & 20.16 & 8.16 \\
OSFTC & 3.79 & 8.66  & 4.59 & 4.18 & {12.06} & {5.59} & 4.34 & 12.86 & 5.89 & 5.52 & {18.02} & {7.76}
\end{tabular}
\caption{Scene flow results on non-occluded pixels for the KITTI test set. SSF also outperforms published methods with two-frame inputs.}
\label{table:sf_noc_results}
\end{table*}

We test our semantic scene flow algorithm (\textbf{SSF}) with 3 iterations of cascaded prediction
on the challenging KITTI 2015 benchmark~\cite{menze2015object}. We evaluate the performance of our
disparity estimates for two frames (\textbf{D1}, \textbf{D2}), flow estimates (\textbf{Fl})
for the reference frame, and scene flow estimates (\textbf{SF})
for foreground pixels (\textbf{fg}),
background pixels (\textbf{bg}), and all pixels (\textbf{all}).
See Table~\ref{table:sf_all_results} for experimental
results on all pixels, and Table~\ref{table:sf_noc_results} for non-occluded pixels.
We evaluate SSF cascades learned to refine PRSF~\cite{vogel2013piecewise} initializations (\textbf{SSF-P}) and also apply the learned parameters to OSF~\cite{menze2015object}
initializations (\textbf{SSF-O}). Our cascaded approach is superior to the published two-frame scene flow algorithms
with respect to all evaluation metrics. SSF-P is about 60\% more accurate than the two-frame PRSF method; SSF-P is overall 2\% more accurate than OSF, while 10 times faster. At the time of submission, the only published work that performed better than our SSF approach were the multi-frame PRSM~\cite{vogel20153d} and OSFTC~\cite{Neoral2017CVWW} methods, which require additional data. The concurrently developed ISF method~\cite{Behl2017ICCV} uses external training data for instance segmentation and disparity estimation, leading to further improvements over our approach at the cost of slower speed. 

\begin{figure*}
\begin{tabular}{cccc}
&&&\\
&&&\\
&&&\\
\subfigimg[width=0.24\linewidth]{Stereo Input}{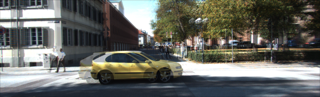} &
\subfigimg[width=0.24\linewidth]{GT Disparity}{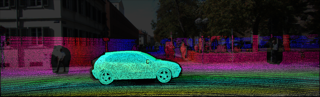} &
\subfigimg[width=0.24\linewidth]{Motion Input}{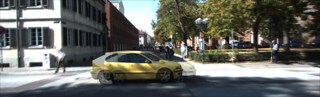} &
\subfigimg[width=0.24\linewidth]{GT Flow}{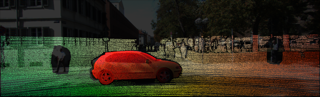} \\
\subfigimg[width=0.24\linewidth]{Disparity Input}{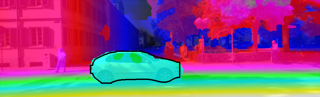} &
\subfigimg[width=0.24\linewidth]{Updated Disparity}{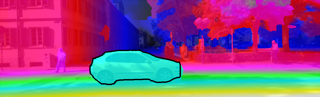} &
\subfigimg[width=0.24\linewidth]{Flow Input}{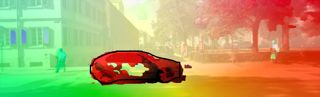} &
\subfigimg[width=0.24\linewidth]{Updated Flow}{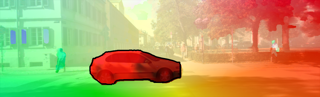} \\
\subfigimg[width=0.24\linewidth]{Initial Disparity Error}{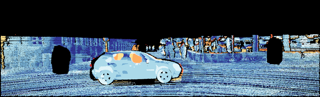} &
\subfigimg[width=0.24\linewidth]{Updated Disparity Error}{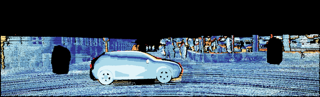} &
\subfigimg[width=0.24\linewidth]{Initial Flow Error}{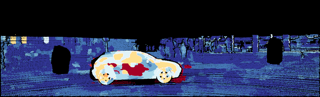} &
\subfigimg[width=0.24\linewidth]{Updated Flow Error}{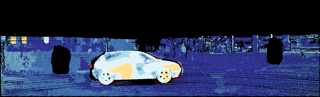} \\
\vspace{-0.7em}&&& \\
\subfigimg[width=0.24\linewidth]{Stereo Input}{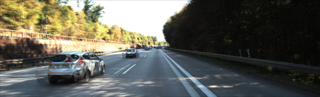} &
\subfigimg[width=0.24\linewidth]{GT Disparity}{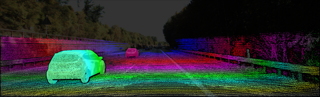} &
\subfigimg[width=0.24\linewidth]{Motion Input}{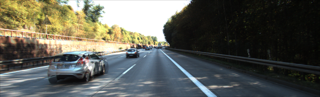} &
\subfigimg[width=0.24\linewidth]{GT Flow}{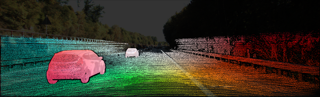} \\
\subfigimg[width=0.24\linewidth]{Disparity Input}{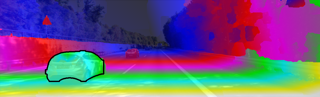} &
\subfigimg[width=0.24\linewidth]{Updated Disparity}{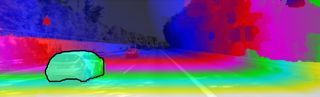} &
\subfigimg[width=0.24\linewidth]{Flow Input}{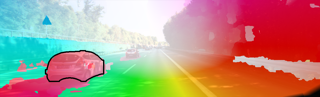} &
\subfigimg[width=0.24\linewidth]{Updated Flow}{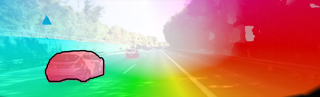} \\
\subfigimg[width=0.24\linewidth]{Initial Disparity Error}{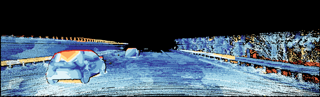} &
\subfigimg[width=0.24\linewidth]{Updated Disparity Error}{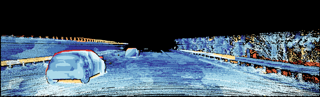} &
\subfigimg[width=0.24\linewidth]{Initial Flow Error}{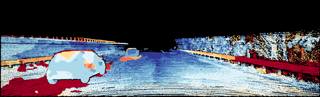} &
\subfigimg[width=0.24\linewidth]{Updated Flow Error}{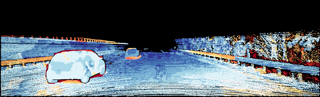} \\
\subfigimg[width=0.24\linewidth]{Stereo Input}{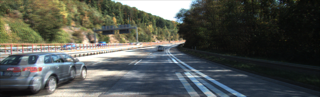} &
\subfigimg[width=0.24\linewidth]{GT Disparity}{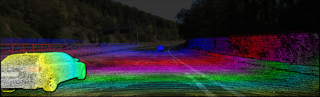} &
\subfigimg[width=0.24\linewidth]{Motion Input}{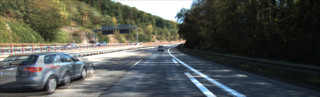} &
\subfigimg[width=0.24\linewidth]{GT Flow}{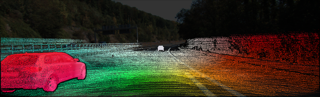} \\
\subfigimg[width=0.24\linewidth]{Disparity Input}{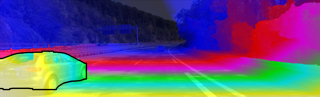} &
\subfigimg[width=0.24\linewidth]{Updated Disparity}{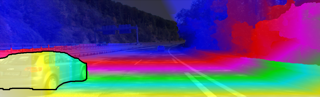} &
\subfigimg[width=0.24\linewidth]{Flow Input}{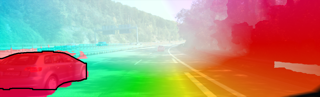} &
\subfigimg[width=0.24\linewidth]{Updated Flow}{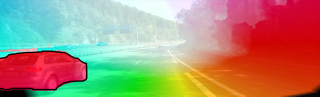} \\
\subfigimg[width=0.24\linewidth]{Initial Disparity Error}{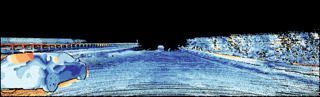} &
\subfigimg[width=0.24\linewidth]{Updated Disparity Error}{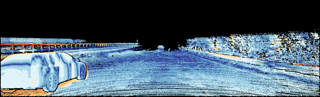} &
\subfigimg[width=0.24\linewidth]{Initial Flow Error}{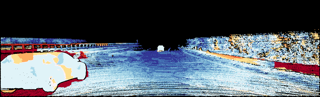} &
\subfigimg[width=0.24\linewidth]{Updated Flow Error}{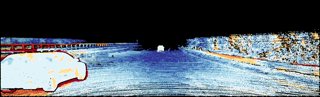} \\
\hline \vspace{-1em}\\
\subfigimg[width=0.24\linewidth]{Stereo Input}{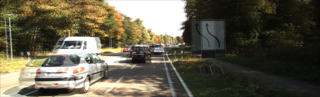} &
\subfigimg[width=0.24\linewidth]{GT Disparity}{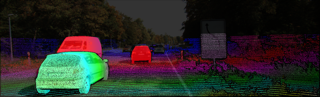} &
\subfigimg[width=0.24\linewidth]{Motion Input}{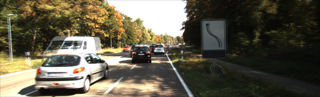} &
\subfigimg[width=0.24\linewidth]{GT Flow}{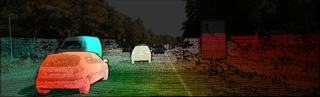} \\
\subfigimg[width=0.24\linewidth]{Disparity Input}{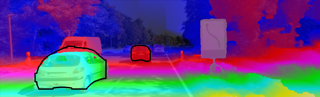} &
\subfigimg[width=0.24\linewidth]{Updated Disparity}{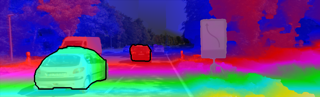} &
\subfigimg[width=0.24\linewidth]{Flow Input}{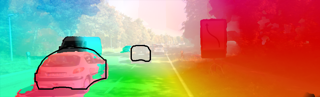} &
\subfigimg[width=0.24\linewidth]{Updated Flow}{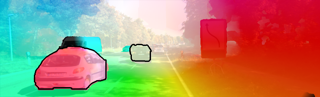} \\
\subfigimg[width=0.24\linewidth]{Initial Disparity Error}{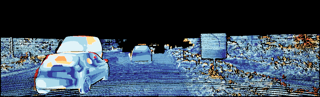} &
\subfigimg[width=0.24\linewidth]{Updated Disparity Error}{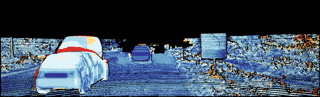} &
\subfigimg[width=0.24\linewidth]{Initial Flow Error}{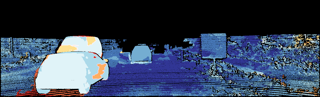} &
\subfigimg[width=0.24\linewidth]{Updated Flow Error}{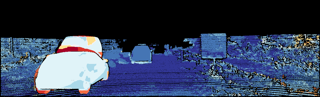} \\
\multicolumn{4}{c}{\hspace{-1em}\includegraphics[width=0.95\linewidth]{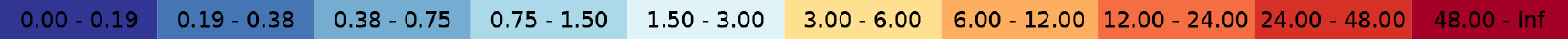}}
\vspace{1.5em}
\end{tabular}
\caption{Visualization of our semantic scene flow estimates for four sequences from the KITTI training set, using a cascade initialized with noisy PRSF estimates. The initial or updated segmentation mask (black lines) is overlaid on the disparity and optical flow estimates.
  The last sequence is a failure case where an imperfect segmentation leads to large disparity and flow errors.}
\label{fig:vis_results}
\end{figure*}

We visualize the qualitative performance of our \textbf{SSF-P} method on training data in Fig.~\ref{fig:vis_results}. 
In Table~\ref{table:sf_training_results}, we evaluate the performance gain provided by each
stage of the cascade on the training set. There is an improvement at the first stage 
of the cascade when modeling segmentation and geometry independently at each frame, followed by another
improvement at the second stage when temporal consistency is introduced. At the
third stage, performance starts to saturate.

\setlength\tabcolsep{3pt} 

\vspace*{-14pt}
\paragraph{Speed}
Scene flow estimation is a computationally demanding task,
and efficient algorithms~\cite{lv2016continuous}
usually sacrifice accuracy for speed.   
Although the number of variables in our scene flow representation is huge and we make 
pixel-level predictions, our cascaded algorithm with MATLAB/C++ implementation on a single-core 2.5 Ghz CPU remains efficient. The main reason is
that we disentangle the output space,
and utilize efficient message-passing algorithms~\cite{felzenszwalb2006efficient,chen2016full} to solve each high-dimensional inference problem.
Most of the computation time is spent on feature evaluation,
and could be accelerated using parallelization.

\begin{table}
\centering
\begin{tabular}{c | c c  c  c  c }
{\bf}&  {\bf Seg} & {\bf D1-fg}  & {\bf D1-bg}&  {\bf Fl-fg}&  {\bf Fl-bg} \\ \hline
 PRSF   &10.1 & 5.00 & 3.27 & 8.54 & 4.04 \\
 Iter 1 &9.79 & 3.69 & 3.18 & 8.38 & 3.67 \\
 Iter 2 &8.41 & 3.50 & 3.15 & 8.20 & 3.65\\
 Iter 3 &8.40 & 3.49 & 3.15 & 8.20 & 3.65\\
 GT Seg & 0   & 2.19 & 3.05 & 7.61 & 3.56
\end{tabular}
\caption{Results of SSF-P on the KITTI validation set.  Each stage of the cascade makes further improvements to the noisy PRSF initialization.
In the last row, we show that when given a perfect segmentation mask,
predictions are improved by a large margin.}
\label{table:sf_training_results}
\end{table}

\vspace*{-14pt}
\paragraph{Failure Cases}
As shown in Fig.~\ref{fig:vis_results}, in challenging cases where the semantic segmentation algorithm fails to detect vehicle boundaries, our scene flow estimates can be inaccurate.
As previously studied for semantic optical flow methods~\cite{bai2016exploiting}, we conducted an experiment using ground truth segmentation masks and witnessed a significant performance gain; see Table~\ref{table:sf_training_results}. 
Our framework for cascaded scene flow estimation will immediately benefit from future advances in semantic instance segmentation.

\vspace*{-4pt}
\section{Conclusion}
\vspace*{-4pt}

In this paper, we utilize semantic cues to identify rigidly moving objects, and thereby produce more accurate scene flow estimates for real-world scenes.
Our cascaded prediction framework allows computationally efficient recovery of high-dimensional motion and geometry estimates, and can flexibly utilize cues from sophisticated semantic segmentation algorithms.
We improve on the state-of-the-art for the challenging KITTI scene flow benchmark~\cite{menze2015object,Geiger:2012:We}.
While our experiments have focused on using vehicle detections to improve scene flow estimates for autonomous driving, our cascaded scene flow framework is directly applicable to any category of objects with near-rigid motion.

\paragraph{Acknowledgements} This research supported in part by ONR Award Number N00014-17-1-2094. Some work was completed by Zhile Ren during an internship at NVIDIA.

\cleardoublepage
{\small
\bibliographystyle{ieee}
\bibliography{iccv2017_ssf}
}

\end{document}